\RequirePackage{fix-cm}
\documentclass[twocolumn]{svjour3}          % twocolumn
\smartqed  % flush right qed marks, e.g. at end of proof
\makeatletter
\expandafter\let\csname opt@amsmath.sty\endcsname\relax
\makeatother

\usepackage{graphicx}
\usepackage{gensymb}
\usepackage{footnote}
\usepackage{xcolor}
\usepackage{hyperref}
\usepackage{caption}
\usepackage{amsmath}
\usepackage{wrapfig}
\usepackage[utf8x]{inputenc}
\usepackage[switch]{lineno}
\begin{document}

\onecolumn
\begin{center}
\vspace{30pt}
\textit{International Journal on Document Analysis and Recognition (IJDAR)}\\
\vspace{30pt}
\textbf{MRZ code extraction from visa and passport documents using convolutional neural network}\\
This is an preprint of https://doi.org/10.1007/s10032-021-00384-2 \\
\vspace{30pt}
Yichuan Liu*, Hailey James, Otkrist Gupta and Dan Raviv\\
  \vspace{5pt}
Lendbuzz \\
  \vspace{5pt}
 {yichuan}.{liu}@lendbuzz.com \\
  \vspace{5pt}
125 High Street, Boston, Massachusetts \\
  \vspace{5pt}
*corresponding author\\
\end{center}

\newpage
\twocolumn

\title{MRZ code extraction from visa and passport documents using convolutional neural networks%\thanks{Grants or other notes\noalign{\smallskip}\hline\noalign{\smallskip}
%about the article that should go on the front page should be
%placed here. General acknowledgments should be placed at the end of the article.}
}

%\titlerunning{Short form of title}        % if too long for running head

\author{Yichuan Liu         \and
        Hailey James \and
        Otkrist Gupta \and
        Dan Raviv%etc.
}

%\authorrunning{Short form of author list} % if too long for running head

\institute{Yichuan Liu \at
              125 High Street, Boston, Massachusetts \\
              \email{yichuan.liu@lendbuzz.com}           %  \\
%             \emph{Present address:} of F. Author  %  if needed
           \and
           Dan Raviv \at
              \email{dan.raviv@lendbuzz.com}
}

\date{Received: date / Accepted: date}
% The correct dates will be entered by the editor

\maketitle

\begin{abstract}
Detecting and extracting information from the Machine-Readable Zone (MRZ) on passports and visas is becoming increasingly important for verifying document authenticity. However, computer vision methods for performing similar tasks, such as optical character recognition (OCR), fail to extract the MRZ from digital images of passports with reasonable accuracy. We present a specially designed model based on convolutional neural networks that is able to successfully extract MRZ information from digital images of passports of arbitrary orientation and size. Our model achieves 100\% MRZ detection rate and 99.25\% character recognition macro-f1 score on a passport and visa dataset.

% \PACS{PACS code1 \and PACS code2 \and more}
% \subclass{MSC code1 \and MSC code2 \and more}
\end{abstract}

\section{Introduction}
\label{sec:1}
In domains such as finance, immigration and administration, digital copies of passports are playing an increasingly important role in identity and information verification and fraud detection. However, automatic information retrieval from passports and visas can be difficult due to non-uniform passport and visa layouts.  Information such as name, birth date, expiration date, and issue date appear in a variety of formats and locations on passports and visas from different issuing authorities. Additionally, unlike physical passports and visas which can be examined for authenticity, digital copies of these documents present a lower barrier to forgery and manipulation. Simple image editing software can be used to alter key details on the passport or visa for purposes of fraud. 

The Machine-Readable Zone (MRZ) on passports and visas is critical for combating both of these challenges. For purposes of information verification, the MRZ presents key information in a pre-specified format and location. As a fraud example, some of our customers uploaded the passport image they found on the internet instead of their own documents. Being able to read the MRZ and compare it with the information entered by  our customers thus serves as an important step in fraud detection. Similarly, the format of the MRZ makes it more difficult to manipulate than the rest of the passport, requiring domain knowledge and greater attention to detail. Locating and extracting passport and visa MRZ thus presents an important and unique application for computer vision.

We propose a novel neural network model designed specifically for handling MRZ text, with characteristics designed to overcome the challenges unique to MRZ extraction. Specifically, we design an end-to-end trainable MRZ detector and extractor using MobileNetV2 as the backbone and added atrous spatial pyramid pooling layers to enhance receptive fields. For better handling of passport images of various sizes, we propose a novel system in which the first ''coarse'' model extracts the MRZ bounding box and the second ''fine'' model refines bounding box prediction and extracts the MRZ text. This system design offers the additional benefit of decreasing the memory and time required for detection. Our proposed system results in 100\% MRZ detection rate and 99.25\% character recognition macro-f1 score on digital images of passports and visas.

\section{Background}
\subsection{Machine-Readable Zone (MRZ)}
The Machine-Readable Zone (MRZ) appears on passports and visas of most countries to facilitate robust data extraction and processing. Because passports from different states vary in script, style, and format, the MRZ provides a simple way to extract key details from the passport, including the name, passport number, nationality, date of birth, sex, and passport expiration date. Generally appearing near the beginning of the passport on the identity page, the MRZ text typically appears as two 44-character lines on the bottom of the page. While alternative MRZ formats are employed in documents such as ID card and visa issued by some countries, we limit this work to consider only this MRZ format most commonly found in passport and US visa images as typically uploaded by our customers. The MRZ consists only of the Arabic numerals (digits 0-9), the capital letters of the Latin alphabet ('A','B','C',...), and the filler character '\textless'.
While historically used for quickly extracting the important information from a variety of passports, the MRZ is becoming  useful for document verification and manipulation detection. For example, businesses and states can verify that the information encoded in the MRZ matches the information in the visual zone (VZ) of the passport. While highly-motivated and skilled forgers can additionally alter the MRZ, validation of the MRZ information is a simple, low-cost method for detecting basic manipulations, such as name, expiration, or birth date changes. As photographs of passports  gain popularity as a method for verifying identity, accurately and quickly extracting the passport MRZ becomes an essential part of the identity verification pipeline.

\begin{figure*}
    % Use the relevant command to insert your figure file.
    % For example, with the graphicx package use
    \includegraphics[width=1\textwidth]{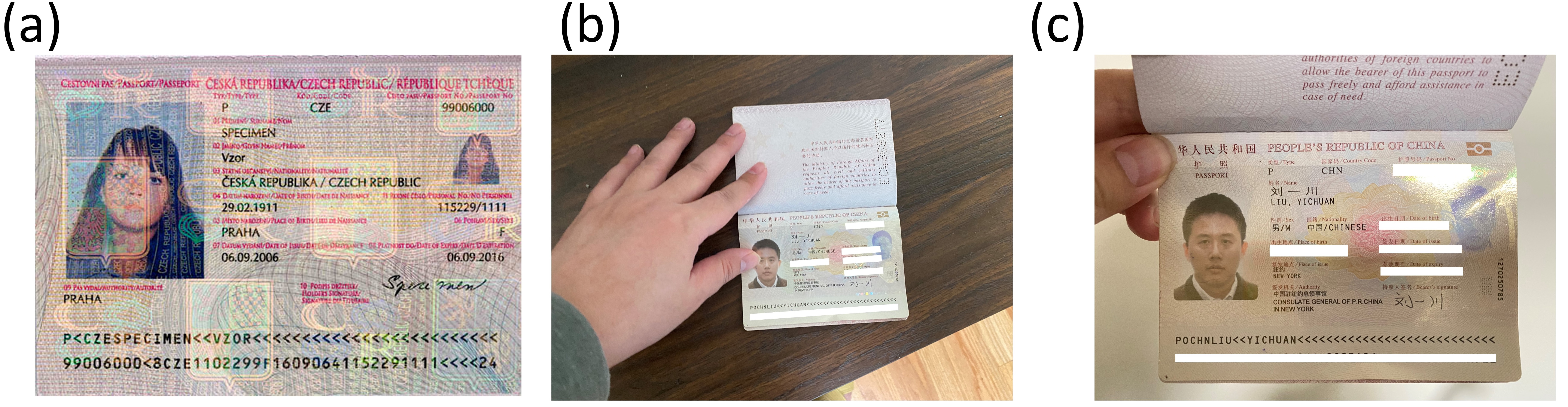}
    % figure caption is below the figure
    \caption{Example passport images. (a) a typical page of the passport contains the 2-line MRZ zone (bottom). The passport page can either occupy only a small part of the image (b) or span the whole image (c).}
    \label{fig:1}       % Give a unique label
\end{figure*}

\begin{figure*}
% Use the relevant command to insert your figure file.
% For example, with the graphicx package use
  \includegraphics[width=1\textwidth]{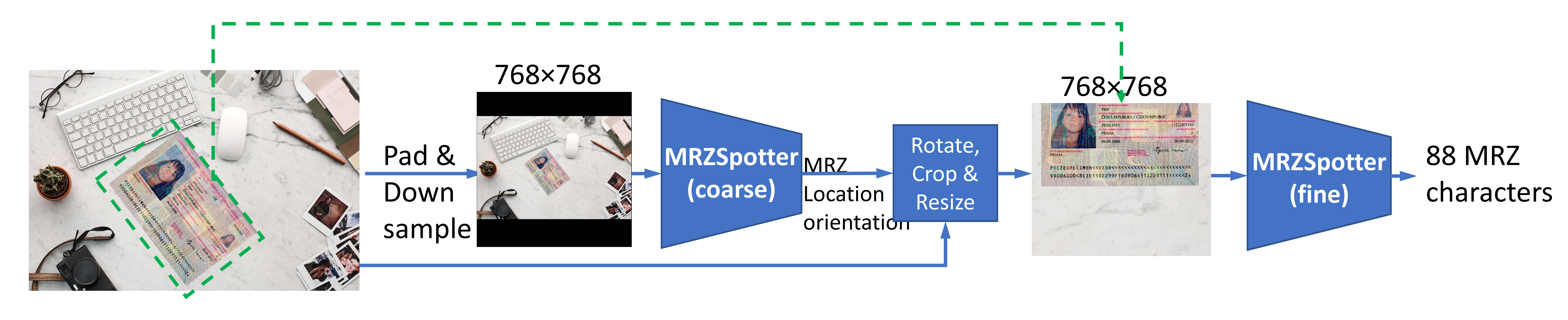}
% figure caption is below the figure
\caption{Overall structure of MRZNet. MRZSpotter (coarse) roughly locates the MRZ region from a down-sampled image whereas MRZSpotter (fine) refines the localization on the original high resolution image and recognizes the MRZ text.}
\label{fig:2}       % Give a unique label
\end{figure*}

\section{Related Work}

Models leveraging advances in deep learning, such as convolutional neural networks (CNNs), have been successfully employed in similar tasks, such as determining the region of interest (ROI) of a photograph \cite{long2015fully}  and optical character recognition (OCR) \cite{wang2017gated}. Among these, MRZ extraction from digital passport images is most related to work in detecting and extracting text in natural scenes. 

\subsection{Text Detection in Natural Scenes}

Several techniques in computer vision have been developed or leveraged for improved performance in text scene detection. \cite{liu_robust_nodate} and \cite{zhu2007} used image binarization to segment text regions. \cite{merino-gracia_head-mounted_2012} \cite{donoser_detecting_2007} and \cite{Gonzlez2012LocationIC} used Maximally Stable Extremal Region (MSER) to improve detection. \cite{fabrizio} and \cite{minetto_text_2011} used morphological operations to segment text regions. \cite{mishra_scene_2012} and \cite{pan_hybrid_2011} used Histogram of Oriented Gradients (HOG) for improved performance.\cite{kasar_multi-script_2012} and \cite{neumann_real-time_2012} used color properties to both detect and extract text regions. 

More recently, the ICDAR 2015 Robust Reading Competition dataset has provided a valuable benchmark for scene text detection and extraction \cite{karatzas_icdar_2015}. Many recent works demonstrated impressive performance on this dataset. \cite{zhang_multi-oriented_2016} use Fully Convolutional Network (FCN) models, trained separately to predict the saliency map of the text regions and the centroid of each character. \cite{shi_detecting_2017} similarly trained a FCN, but on the segments and links of the text which are combined for final detection. \cite{he_single_2017} proposed a single shot attention-based mechanism that attempts a coarse to fine approach to text detection. \cite{hu_wordsup_2017} leveraged a weakly supervised framework that uses word annotations to train the character detector. \cite{zhou2017} proposed EAST (An Efficient and Accurate Scene Text Detector), which skips intermediate steps like candidate aggregation and word partitioning to directly predict words and text lines. \cite{long_textsnake_2018} attempted to consider more free-form text examples such as curved text using a FCN to estimate geometric attributes of the scene regions.\cite{liao_textboxes_2018} iterated on the object detector method proposed in \cite{Liu_2016}. \cite{deng_pixellink_2018} proposed a novel method for scene text detection using instance segmentation. \cite{lyu_multi-oriented_2018} and \cite{deng_detecting_2019} leveraged the corner points of the text bounding boxes for better segmentation and detection. \cite{wang_shape_2019-1} generated different kernel scales for each text instance in order to split close text instances. \cite{dai_fused_2018} combined multi-level features during feature extraction for improved performance. \cite{lyu_mask_2018} improved performance with an architecture inspired by Mask R-CNN \cite{he_mask_2018}. \cite{liao_real-time_2019} proposed a module to perform binarization in a segmentation network. \cite{liu_fots_2018} trained a network for simultaneous detection and recognition by sharing convolutional features between the two processes. \cite{xu_geometry_2019} leveraged multiple branches to achieve geometry normalization. \cite{liu_exploring_2020} built on previous work and incorporate a method to discretize the potential quadrilaterals into various horizontal and vertical positions. So far, \cite{xing_convolutional_2019} achieved the most impressive performance by training on synthetic data, using characters as the basic element, and eliminating ROI pooling.

\subsection{Passport MRZ Detection and Extraction}

While Optical Character Recognition (OCR) based-methods may extract text with reasonably good accuracy, state of the art methods struggle to accurately extract MRZ text. This is evidenced by the relatively poor MRZ detection rate of PassportEye \cite{passporteye} which is based on Tesseract OCR \cite{smith2007overview}.  Similarly, models designed for scene text extraction are not naturally well-suited for MRZ extraction. For example, end-to-end scene text recognition models such as FOTS \cite{liu2018fots} and Mask Textspotter \cite{lyu2018mask} may be able to detect and recognize the MRZ. However, these models are designed to handle text lines with arbitrary number of characters and employed techniques such as LSTM \cite{hochreiter1997long} to recognize text. Since common MRZ text found in passport and visa is always 2 lines, 44 characters per line, a specifically designed neural network architecture will likely improve performance. Additionally, typical passport images used for identity verification purposes are taken with a smartphone, resulting in a high resolution images in which the passport appears in various places and at various sizes (see Figure ~\ref{fig:1}), presenting an additional challenge.

In 2011, \cite{bessmeltsev_high-speed_2011} presented a hardware-based method for portable passport readers for detecting and reading the MRZ of physical passports. \cite{lee_character_2015} proposed a method for extracting the passport MRZ using template matching, but only for images in which the passport is surrounded by a black border. \cite{chernyshova_optical_2019} explored optical font recognition for forgery detection in passport MRZs. \cite{petrova_methods_2019} discussed methods for correcting or post-processing passport MRZ recognition results. \cite{hartl2014} presented an algorithm for reading MRZ images on mobile devices, achieving an MRZ detection rate of 88.18\% with 5 frames and 56.1\% with single frame, along with a character reading rate of 98.58\%. In comparison our model boasts a 100\% single frame MRZ detection rate and 99.25\% character recognition macro-f1 score on passport and visa images.

\section{Methodology}
\label{method}
MRZNet is a framework that detects and recognizes the MRZ text in images of passports and visas given arbitrary orientation and sizes. This section describes the details of the architecture of MRZNet.

\subsection{Overall architecture}
\label{overall_architecutre}

The overall architecture of MRZNet is illustrated in Figure ~\ref{fig:2}. It includes two sub neural networks, MRZSpotter (coarse) and MRZSpotter (fine), which share similar architectures. The high resolution original image is first padded to be a square and then down-sampled to 768 x 768 as input to MRZSpotter (coarse). MRZSpotter (coarse) localizes the MRZ region and outputs the bounding box location and orientation.  We then rotate the original image to make it upright, crop the image centered at the  bounding box center and pad/resize the image accordingly to obtain a 768 x 768 image in which the MRZ region is roughly placed in the center and spans the whole image. This image is then fed into MRZSpotter (fine) for finer localization and MRZ code recognition. We adopt this architecture for handing passport/visa images of arbitrary orientation and sizes. A real world passport/visa image, whether it is scanned or taken from a smart phone, is usually of high resolution. Depending on how the user captures the image, the MRZ region can either occupy only a small region of the image (Figure ~\ref{fig:1}(b)) or it may span the whole image (Figure ~\ref{fig:1} (c)). Feeding the high resolution image directly into a neural network is not only time and memory consuming but may result in poor MRZ code recognition results for images like Figure ~\ref{fig:1} (c). Specifically, localizing these high resolution images would demand a very large receptive field within the neural network, increasing time and memory requirements On the other hand, feeding a down-sampled image to a neural network would result in poor MRZ recognition results for images like Figure ~\ref{fig:1} (b) because the text will be unrecognizable in low resolution. To solve this dilemma, we propose an architecture that first roughly localizes the MRZ region using a down-sampled image, standardizes the images (see Figure ~\ref{fig:1} (c))  and finally performs MRZ text recognition.

\subsection{MRZSpotter}
\label{MRZSpotter}
\begin{figure*}
% Use the relevant command to insert your figure file.
% For example, with the graphicx package use
  \includegraphics[width=1\textwidth]{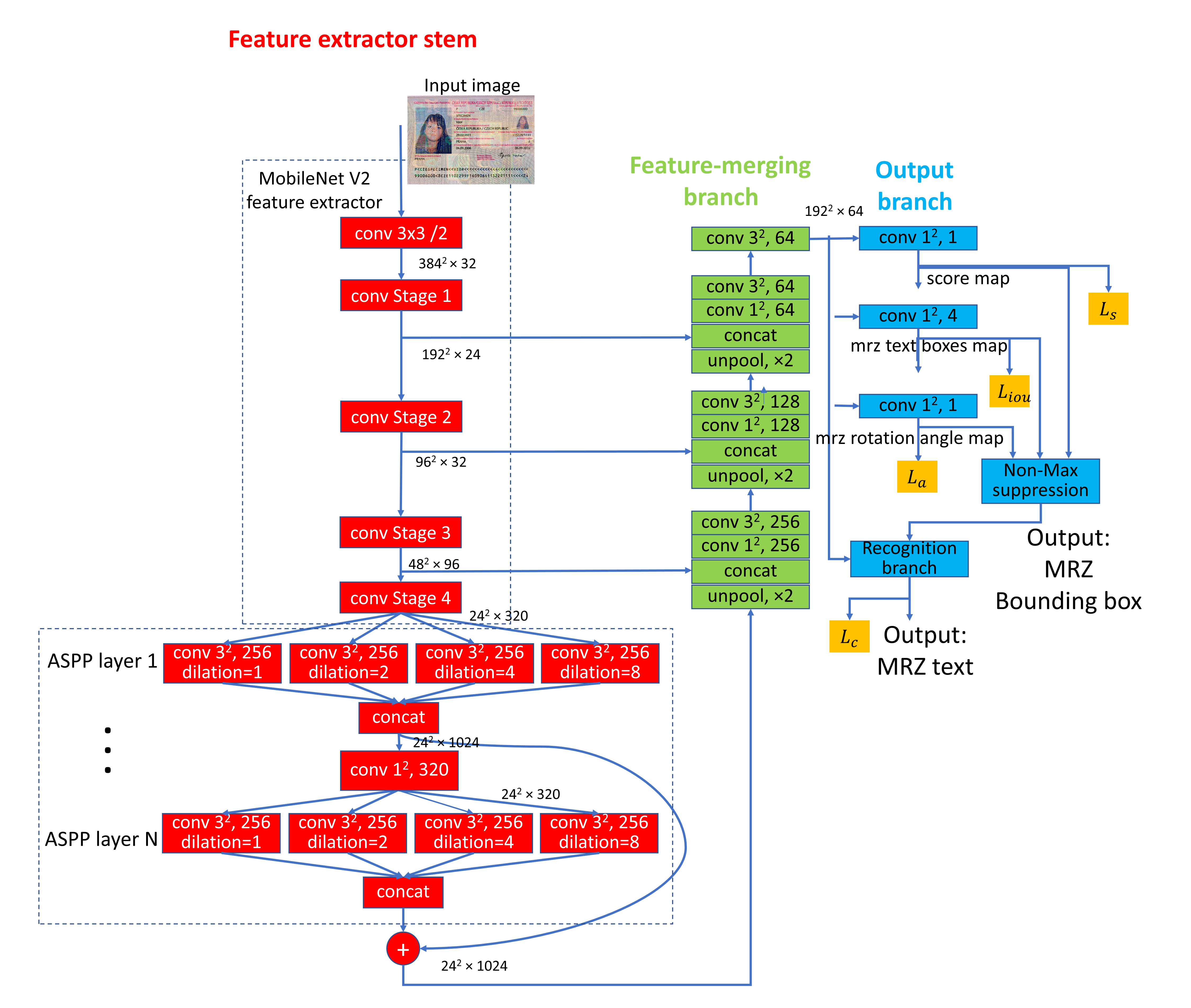}
% figure caption is below the figure
\caption{MRZSpotter with N atrous spatial pyramid pooling (ASPP) layers. Both MRZSpotter (coarse) and MRZSpotter (fine) used the same architecture and loss as shown in this figure, though with different parameter N. We stacked N ASPP layers on top of the last convolutional stage of MobileNetV2 to increase the receptive field and add a text recognition branch in addition to text localization branch.}
\label{fig:3}       % Give a unique label
\end{figure*}

    The architecture of MRZSpotter is shown in Figure ~\ref{fig:3}. Because we use CPU to run models in production, we adopt MobileNetV2 \cite{sandler2018} as the backbone to reduce computational cost. Similar to EAST, we concatenate up-sampled high-level semantic feature maps with low-level feature maps and merge them gradually in a U-shaped architecture. This way the neural network utilizes the features from different levels and will be able to detect MRZ regions of different sizes. In some examples, the line of text will span the whole image (see Figure ~\ref{fig:1} (c)). For better handling of these images, a larger receptive field is required to look at the "big picture'' of the image in order to accurately detect the large text bounding box. We applied atrous spatial pyramid pooling (ASPP) at the end of the MobileNetV2 feature extractor to accommodate these larger receptive fields. ASPP have been previously adopted by \cite{sermanet2013}, \cite{giusti2013}, \cite{simonyan2014} and \cite{chen2017} for field-of-view enlargement. To further increase the field-of-view, we stacked multiple layers of ASPP as demonstrated in ResNet \cite{he2016}. After feature-merging, 1x1 convolutional layers are applied to the output to determine the likelihood that an MRZ region is present in the pixel (the score map), the location of MRZ text boxes (4 channels, distance of the pixel locations to the top, right, bottom and left boundaries of the rectangle, respectively) and the MRZ box rotation angle. The non-maximum suppression algorithm is applied to select the most probable MRZ bounding box. Finally, a recognition branch is applied to the MRZ bounding box and the output map of the feature-merging branch to extract the MRZ text.

\subsection{MRZSpotter pipeline}
We first extract feature maps from a passport/visa image using a MobileNetV2 backbone. At the end of the stage 4 convolutional layers, MobileNetV2 produces 320 feature maps of size $24 \times 24$. We then add four convolutional layers that run in parallel to form a ASPP layer. These four convolutional layers have a dilation rate \cite{chen2017} of 1, 2, 4 and 8. We concatenate the feature maps produced by these four layers (the concatenation layer) and then applied a $1 \times 1$ convolutional layer to reduce the number of feature maps to 320 before feeding the resulting feature maps to the next ASPP layer. Shortcuts were added between the concatenation layer of two ASPP layers similar to ResNet. After N ASPP layers, we bilinearly up-sample (un-pool) the feature maps to size $48 \times 48$ before concatenate them with the feature map outputs from the end of stage 3 convolutional layer of MobileNetV2. A $1 \times 1$ followed by a $3 \times 3$ convolutional layer is used to fuse these feature maps. We then bilinearly up-sample the resulting feature maps to $96 \times 96$ sizes and concatenate them with output of the stage 2 convolutional layers of MobileNetv2.  After fusing the feature maps with 2 convolutional layers, we bilinearly upsampled them to $192 \times 192$ sizes and concatenate them with the output of stage 1 convolutional layers of MobileNet V2. Three convolutional layers are then applied to fuse and extract features from these feature maps to produce the output of feature-merging branch, which is composed of 64 feature maps of size $192 \times 192$. Similar to EAST\cite{zhou2017}, for each pixel in the output of feature-merging branch, we apply $1 \times 1$ convolutional layers at the output branch to produce a 0-1 probability score which indicates the presence of MRZ (the score map) at the pixel, the distance from the top, bottom, left and right of the mrz bounding box to the pixel (MRZ text box map) and the rotation angle of the bounding box (mrz rotation angle map). Because we have $192 \times 192$ pixels, a total of $192 \times 192 = 36864$ bounding boxes are produced as a result. We reject those bounding boxes that have a probability score lower than 0.5 and use non-max suppression (NMS) to fuse the rest of the bounding boxes. The bound box that has the highest score is then selected as input to the recognition branch.

\subsection{Recognition branch}
\label{recognition_branch}
Both the MRZSpotter (coarse) and MRZSpotter (fine) include a recognition branch for recognizing MRZ text. Our recognition branch is inspired from \cite{he2018}. Figure \ref{fig:4} shows the architecture. Given the quadrilateral MRZ region from NMS, we sample a 16 by 352 grid from the convolutional map at the output of feature-merging branch. Similarly to \cite{he2018}, we used bilinear sampling. More specifically, the feature vector $v_p$ of a sampling point $p$ at spatial location $(p_x, p_y)$, is calculated as follows:
\begin{equation}
v_p=\sum_{i=0}^{3}v_{pi}g(p_x, p_{ix})g(p_y, p_{iy})
\end{equation}
where $v_{pi}$ refer to the surrounding four points of point $p$ and $g(p_1, p_2)$ refers to the bilinear interpolation function.

After extracting the sampling grid, three layers of 3x3 convolution and 2x2 max-pooling are applied to down sample the extracted feature map from 16x352 (points) to 2x44 (lines by characters per line). We doubled the number of channels with each down-sampling. Finally, a 1x1 convolutional layer is applied to reduce the number of channels to 37 (the number of valid characters in MRZ code) and softmax is applied to obtain the probability of occurrence for each of the 88 characters.

\begin{center}
    % Use the relevant command to insert your figure file.
    % For example, with the graphicx package use
    \includegraphics[width=0.4\textwidth]{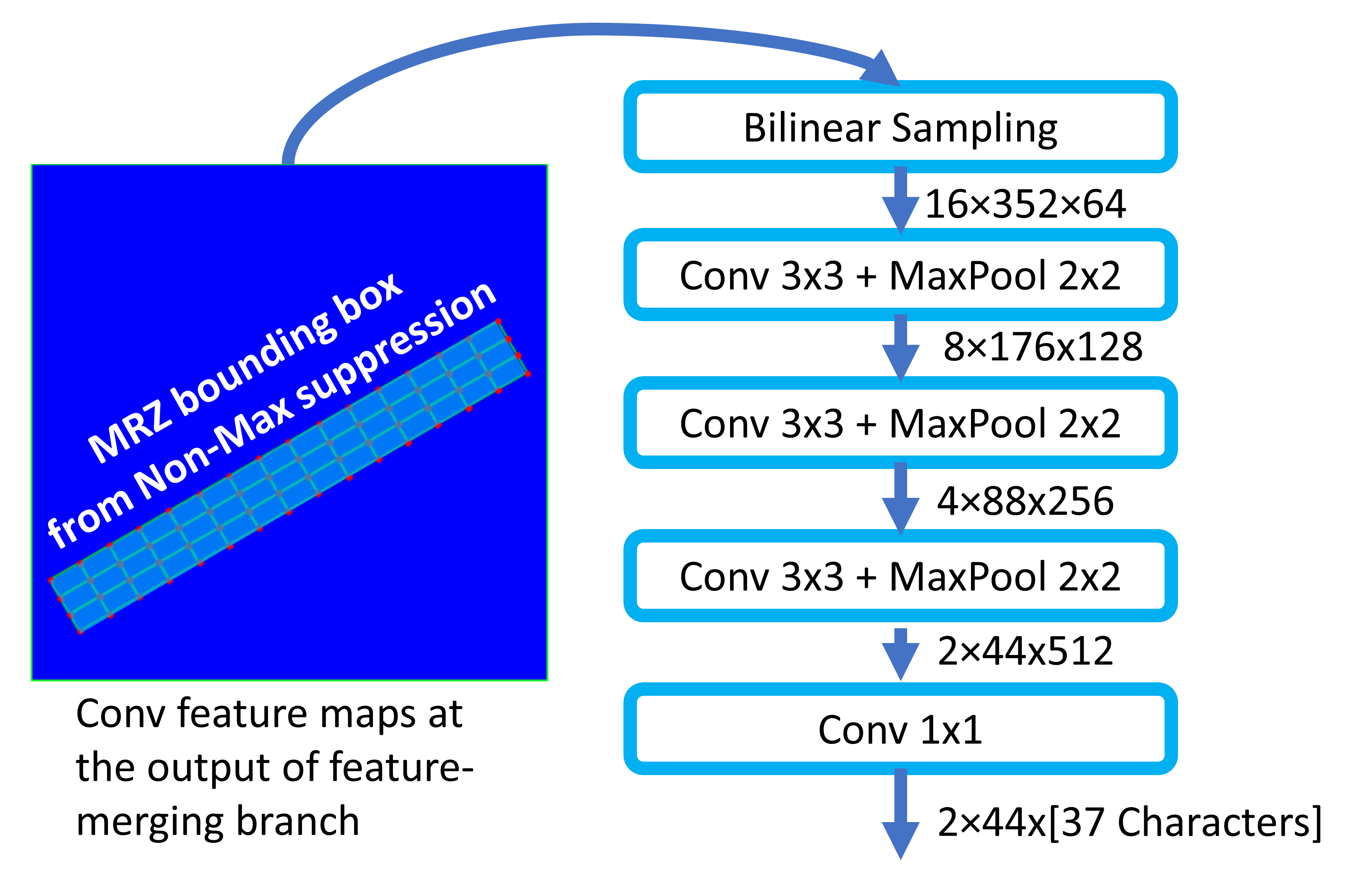}
    % figure caption is below the figure
    \captionof{figure}{Architecture of the recognition branch. Instead of adopting LSTM for recognizing arbitrary length text as in typical scene text recognition networks, we bilinearly sample a 16 by 352 grid from the output convolutional layer of EAST and pass it through several full convolutional layers and max pooling layers to reduce the feature map size to 2 by 44 for recognizing the 2 lines, 44 characters per line MRZ text.}
    \label{fig:4}       % Give a unique label
\end{center}

\subsection{Loss functions}
\label{details}
MRZSpotter (coarse) and MRZSPotter (fine) are trained separately with the same loss function:
\begin{equation}
L=L_{s}+\lambda_{g}L_{g}+\lambda_{c}L_{c}
\end{equation}
where $L_{s}$ is the loss for score map, $L_{g}$ is the loss for geometry, $L_{c}$ is the loss for character classification. In our experiment, we set $\lambda_{g}$ and $\lambda_{c}$ to be 1.
For the loss of score map, we used dice loss instead of the balanced cross-entropy loss as adopted by EAST \cite{zhou2017} due to its higher performance as reported in \cite{east1}:
\begin{equation}
L_{s}=1-\frac{2\sum_{x}s_{x}s_{x}^{*}}{\sum_{x}s_{x}+\sum_{x}s_{x}^{*}}
\end{equation}
where $s_{x}$ and $s_{x}^{*}$ are predicted score and ground truth score, respectively. For the geometry loss, we adopt the intersection over union (IoU) loss and rotation angle loss as in EAST \cite{zhou2017}:

\begin{equation}
L_{g} = L_{iou} + \lambda_{a}L_{a}
\end{equation}

\begin{equation}
L_{iou} = \frac{1}{\Omega}\sum_{x\in\Omega}IoU(R_{x}, R_{x}^{*}) 
\end{equation}

\begin{equation}
L_{a} =  (1-cos(\theta_{x}, \theta_{x}^{*}))
\end{equation}
where $R_{x}$, $R_{x}^{*}$, $\theta_{x}$ and  $\theta_{x}^{*}$ are predicted bounding box, ground truth bounding box, predicted orientation and ground truth orientation, respectively. $IoU$ is calculated as follows:
\begin{equation}
IoU(R_{x}, R_{x}^{*}) = \frac{R_{x} \cap R_{x}^{*}}{R_{x} \cup R_{x}^{*}}
\end{equation}
In our experiment, the weight $\lambda_{a}$ is set to 10. For the character classification loss, we used the cross-entropy loss:
\begin{equation}
L_{c} = \sum_{i=0}^{c}y_{i}log(f_{i}(x))
\end{equation}
where $c$ is the number of possible different characters, $f_{i}(x)$ is the network output of class $i$ for image sample $x$, $y_{i}$ is the one hot ground truth label.

\subsection{Implementation details}
\label{details}
For the MobileNetV2 backbone, we loaded weights pre-trained on the ImageNet dataset \cite{deng2009} before fine-tuning on our MRZ dataset. For training MRZSpotter (coarse), we augmented the dataset by randomly rotating the images in the range of [-180\degree, 180\degree] and randomly padding with black borders so that the new image height is in the range of 1-2 times the height of the original image. We additionally applied random cropping image with the constraint of keeping the MRZ region intact. For training MRZSpotter (fine), we augmented the dataset by randomly rotating the image in the range of [-20\degree, 20\degree] with respect to the upright position. We found that it is important to make sure that the rotation angle is small for MRZSpotter (fine). We then cropped the images so that the cropped region is a square and the MRZ region is roughly centered within image with the left and right borders randomly selected to be 0.05-0.25 times the width of the MRZ region. For both MRZSpotter (coarse) and MRZSpotter (fine), we trained the model for 120 epochs with Adam optimizer \cite{kingma2014} and a initial learning rate of 0.0001, $\beta_{1}=0.9$ and $\beta_{2}=0.999$. The learning rate was decreased by a factor of 10 at 60 epochs. The batch size was selected to be 6. The models were trained with a single GeForce RTX 2070 graphic card. The time to train both MRZSpotter (coarse) and MRZSpotter (fine) are approximately 1 day, making it a total of 2 days to train the entire model. Taken together MRZSpotter (fine) and MRZSpotter (coarse), MRZNet has 28.9M parameters. As a comparision, similar deep learning approaches CharNet \cite{xing_convolutional_2019} and FOTS \cite{liu2018fots} has 89.2M and 35.0M parameters, respectively.

\section{Experimental evaluation}
In this section, we evaluate the performance of MRZNet. We also report results from ablation studies to explore the impact of our design choices. 
\label{experiment}
\subsection{Dataset}
We evaluated our algorithm on a dataset consisting of 4820 passport/visa images. The dataset includes 2687 passport images and 2133 visa images from 85 issuing countries. The visa images are all US issued visa the MRZ of which is 2 lines, 44 characters per line. Table \ref{tab:1} summarizes the data distribution over different countries. The dataset contain real world images that were either scanned by a scanner or taken using a smartphone camera and uploaded to our database by our customers.  It may contain perspective distortions, scaling, illumination and resolution variation, even motion blur. It reflects the real world images we encounter on a daily basis. Each of these images contains a single passport or visa. We manually annotated the ground truth bounding box and MRZ text for all images using VGG image annotator \cite{via}. It is common that text documents other than passport and visa appears in an image so training with this dataset allow our model to ignore none MRZ text regions. We used 3482 for training, 723 for validation and 615 for testing. We also tested our approach on the publicly available MRZ dataset MIDV-500 \cite{arlazarov2019midv} and syntheticMRZ \cite{hartl2014}. For MIDV-500, we used the passport images that contain a MRZ zone. We removed images in which MRZ zone is not entirely intact. This resulted in 3335 test images. For syntheticMRZ, we randomly selected 17113 images. For both MIDV-500 and syntheticMRZ, we only include images whose MRZ zone is in the most common format, containing two lines of text, 44 characters each. The file paths to the images for the MIDV-500 and syntheticMRZ dataset is available upon request.

\begin{table}
\centering
% table caption is above the table
\caption{Distribution of our MRZ dataset across issuing countries}
\label{tab:1}       % Give a unique label
% For LaTeX tables use
\begin{minipage}{0.45\textwidth}
\begin{tabular}{lll}
\hline\noalign{\smallskip}
 Country & Num. of samples \\
\noalign{\smallskip}\hline\noalign{\smallskip}
United States & 2194 \\
Brazil & 1114 \\
China & 931 \\
India & 48 \\
Guatemala & 45 \\
Venezuela & 42 \\
Colombia & 33 \\
El Salvador & 31 \\
Mexico & 31 \\
United Kingdom & 29 \\
Ecuador & 20 \\
Australia & 17 \\
Italy & 15 \\
Other & 270 \\
\noalign{\smallskip}\hline
\end{tabular}
\end{minipage}
\end{table}

\subsection{Comparison with existing solution}
\begin{figure*}
    % Use the relevant command to insert your figure file.
    % For example, with the graphicx package use
    \includegraphics[height=0.9\textheight]{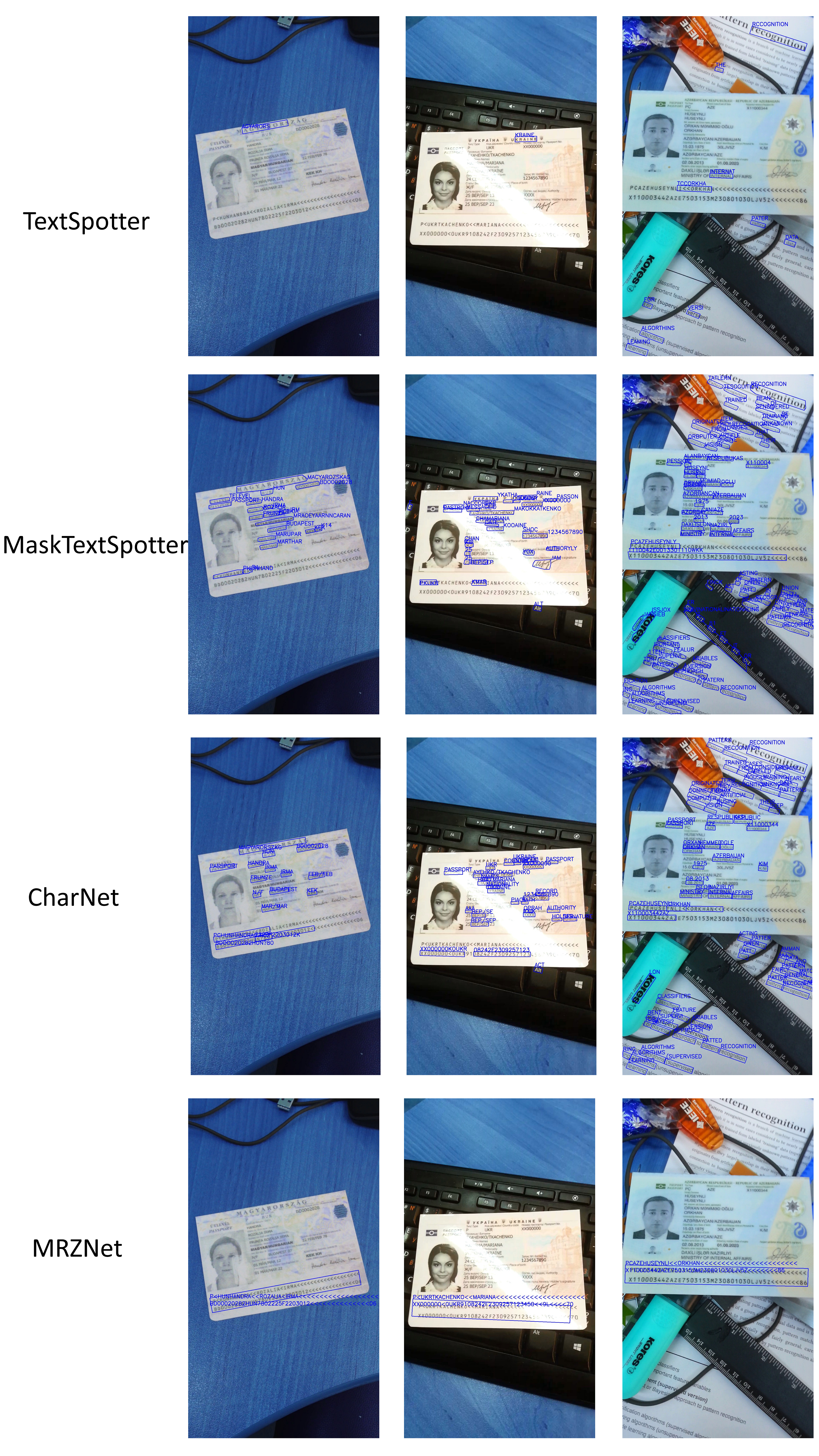}
    % figure caption is below the figure
    \caption{MRZ recognition results by end-to-end deep learning approaches. From top to bottom rows: TextSpotter, MaskTextSpotter, CharNet and MRZNet}
    \label{fig:5}       % Give a unique label
\end{figure*}

We compared our MRZNet against existing MRZ recognition solutions: 1) PassportEye \cite{passporteye} which is based on Tesseract OCR \cite{smith2007overview}, 2) MRZ-Detection \cite{mrzdetection} and 3) UltimateMRZ \cite{ultimatemrz}, a deep learning based commercial solution that is based on LSTM \cite{hochreiter1997long}. We also compared it with end-to-end neural network based text spotting approaches MaskTextSpotter \cite{lyu2018mask}, the EAST based TextSpotter \cite{he2018} and CharNet \cite{xing_convolutional_2019}. For these approaches, we downloaded code and weights trained on ICDAR2015 from official implementations. It can be seen from Table \ref{tab:2}, Table \ref{tab:3} and Table \ref{tab:4}  that our MRZNet outperforms each of the three comparison MRZ detection models as well as other deep learning based end-to-end text retrieval models by a large margin.  In Table \ref{tab:2}, Table \ref{tab:3} and Table \ref{tab:4}, for MRZNet, PassportEye, MRZ-Detection, and UltimateMRZ, the detection rate is defined as the ratio of images whose MRZ character recognition accuracy is higher than 50\%, as the ground truth bounding box of SyntheticMRZ dataset is not available, and because PassportEye and MRZ-detection do not output the predicted bounding box. For the three deep learning based approaches, it is rare that the MRZ character recognition accuracy is higher than 50\% for an image, so we consider the detection is a success if a text box is found in the MRZ region based on the ground truth bounding box. Figure \ref{fig:5} shows examples of text detection results by the various deep learning based approaches. In addition to these approaches, Hartl et al. \cite{hartl2014} achieved character recognition rate of 98.6\% on the SyntheticMRZ dataset. Their MRZ detection rate, however, is only  56.1\% (single frame) and 88.2\% rate (5 frames) whereas our single frame MRZ detection rate is 88.66\% on the SyntheticMRZ dataset. One possible explanation of this large performance gap is that most existing algorithms rely on traditional image processing techniques or the output of a general OCR, while our method employs convolutional neural network as feature extractor for end-to-end detection and recognition. Additionally, MRZNet is specifically designed to handle MRZ detection and recognition which assumes a fixed target of two lines of 44 characters each whereas the end-to-end scene text detectors proposed in the literature are designed for text lines of arbitrary length.  We also reported the run time of all approaches in Table \ref{tab:5}

\begin{table}
\centering
% table caption is above the table
\caption{MRZ detection and character recognition (in macro-f1 score) results on our test set for MRZNet and other solutions}
\label{tab:2}       % Give a unique label
% For LaTeX tables use
\begin{minipage}{0.45\textwidth}
\begin{tabular}{lll}
\hline\noalign{\smallskip}
Method & MRZ Detection & Character Recog. \\
\noalign{\smallskip}\hline\noalign{\smallskip}
PassportEye & 26.50\% & 84.47\%  \\
MRZ-Detection & 52.36\% & 95.01\% \\
UltimateMRZ \footnote{Recognition rate are based on an average of 76 out of 88 characters available in the free version} & 68.78\% & 83.04\%  \\
TextSpotter & 21.79\% & 12.07\% \\
MaskTextSpotter & 69.43\% & 13.72\% \\
CharNet & 74.15\% & 35.53\% \\
MRZNet  & \textbf{100.00\%} & \textbf{99.25\%}  \\
\noalign{\smallskip}\hline
\end{tabular}
\end{minipage}
\end{table}

\begin{table}
\centering
% table caption is above the table
\caption{MRZ detection and character recognition (in macro-f1 score) results on MIDV-500 MRZ dataset  \cite{arlazarov2019midv} for MRZNet and other solutions}
\label{tab:3}       % Give a unique label
% For LaTeX tables use
\begin{minipage}{0.45\textwidth}
\begin{tabular}{lll}
\hline\noalign{\smallskip}
Method & MRZ Detection & Character Recog. \\
\noalign{\smallskip}\hline\noalign{\smallskip}
PassportEye & 27.32\% & 64.93\%  \\
MRZ-Detection & 46.30\% & 76.00\% \\
UltimateMRZ \footnote{Recognition rate are based on an average of 76 out of 88 characters available in the free version} & \textbf{77.15\%} & 71.69\%  \\
TextSpotter & 21.20\% & 13.37\%  \\
MaskTextSpotter & 69.15\% &  19.42\% \\
Charnet & 74.18\% &  28.44\%  \\
MRZNet & 73.94\% & \textbf{85.18\%}  \\
\noalign{\smallskip}\hline
\end{tabular}
\end{minipage}
\end{table}

\begin{table}
\centering
% table caption is above the table
\caption{MRZ detection and character recognition (in macro-f1 score) results on SyntheticMRZ \cite{hartl2014} dataset for MRZNet and other solutions. For TextSpotter,  MaskTextSpotter and CharNet, results can not be generated due to lack of ground truth bounding box label}
\label{tab:4}       % Give a unique label
% For LaTeX tables use
\begin{minipage}{0.45\textwidth}
\begin{tabular}{lll}
\hline\noalign{\smallskip}
Method & MRZ Detection & Character Recog. \\
\noalign{\smallskip}\hline\noalign{\smallskip}
PassportEye & 46.87\% & 84.42\%  \\
MRZ-Detection & 86.06\% & 87.59\% \\
UltimateMRZ \footnote{Recognition rate are based on an average of 76 out of 88 characters available in the free version} & 42.40\% & 78.40\%  \\
TextSpotter & NA & NA \\
MaskTextSpotter & NA & NA \\
CharNet & NA & NA \\
MRZNet & \textbf{88.66\%} & \textbf{90.09\%}  \\
\noalign{\smallskip}\hline
\end{tabular}
\end{minipage}
\end{table}

\begin{table}
\centering
% table caption is above the table
\caption{Recognition speed comparison on our test set, ($mean \pm std$). GPU: a single GeForce RTX 3090; CPU:  Intel(R) Xeon(R) Gold 5220R}
\label{tab:5}       % Give a unique label
% For LaTeX tables use
\begin{minipage}{0.45\textwidth}
\begin{tabular}{lll}
\hline\noalign{\smallskip}
Method & CPU time (s) & GPU time(s)  \\
\noalign{\smallskip}\hline\noalign{\smallskip}
PassportEye & $0.68 \pm 0.45$  &   \\
MRZ-Detection & $2.46 \pm 1.01$ & \\
UltimateMRZ & $0.24 \pm 0.16$ & $0.14 \pm 0.16$   \\
TextSpotter & $23.12 \pm 3.91$ &  $0.90 \pm 0.49$  \\
MaskTextSpotter & $6.74 \pm 1.17$   & $1.58 \pm 0.77$ \\
CharNet & $80.62 \pm 9.53$ & $9.25 \pm 5.34$ \\
MRZNet &  $14.82 \pm 0.97$ & $0.51 \pm 0.78$ \\
\noalign{\smallskip}\hline
\end{tabular}
\end{minipage}
\end{table}

\subsection{Ablation Study}
We performed ablation studies to evaluate the effectiveness of the two stage model and the ASPP layers, with the results are reported in Table \ref{tab:6} and Table \ref{tab:7}. From Table \ref{tab:6}, it can be inferred that using only MRZSpotter (coarse) will result in poor MRZ text recognition accuracy (with 67.87\% macro f1-score as the best result). The primary reason is that the resolution of input image is low. However, we also found that the narrowing and box rotation angle range is an important factor since increasing the data augmentation rotation angle of MRZSpotter (fine) results in a worse accuracy. Including a single ASPP layer to MRZSpotter (coarse) improved the MRZ text recognition performance from 56.56\% to 67.87\%. However, the bounding box localization accuracy decreased from 0.8701 to 0.8508 IoU. Considering the objective of MRZSpotter (coarse) is localization, we haven't included any ASPP layers in the final model for MRZSpotter (coarse). From Table \ref{tab:7}, it can be seen that the MRZ recognition accuracy is much improved by using MRZSpotter (fine) after MRZSpotter (coarse). Including one single layer of ASPP improved the MRZ text recognition accuracy from 98.40\% to 98.91\%. By stacking 3 ASPP layers, the accuracy further improved to 99.21\%. These results demonstrate the impact of the proposed two stage model and the ASPP layers. In Table \ref{tab:8}, we compared the different loss function for score map. From the results, it can be seen that the dice loss as adopted by our approach outperforms the balanced cross-entropy loss as adopted by EAST \cite{zhou2017}.
\begin{table}
\centering
% table caption is above the table
\caption{Results on validation set from MRZSpotter (coarse). We show the variation of bounding box detection IOU and MRZ text recognition macro F1-score with different numbers of ASPP layers}
\label{tab:6}       % Give a unique label
% For LaTeX tables use
\begin{tabular}{lll}
\hline\noalign{\smallskip}
Num. of ASPP & IoU & Macro F1-score \\
\noalign{\smallskip}\hline\noalign{\smallskip}
0 & 0.8701 & 56.56\% \\
1 & 0.8508 & 67.87\% \\
\noalign{\smallskip}\hline
\end{tabular}
\end{table}

\begin{table}
% table caption is above the table
\centering
\caption{Results on validation set from MRZSpotter (fine). We show the variation of bounding box detection IOU and MRZ text recognition macro F1-score with different numbers of ASPP layers}
\label{tab:7}       % Give a unique label
% For LaTeX tables use
\begin{tabular}{lll}
\hline\noalign{\smallskip}
Num. of ASPP & IoU & Macro F1-score \\
\noalign{\smallskip}\hline\noalign{\smallskip}
0 & 0.9071 & 98.40\% \\
1 & 0.9059 & 98.91\% \\
3 & 0.9144 & 99.21\% \\

\noalign{\smallskip}\hline
\end{tabular}
\end{table}
\begin{table}
% table caption is above the table
\centering
\caption{Results on validation set from MRZSpotter (fine). We compare results for balanced cross-entropy loss for score map as adopted by EAST \cite{zhou2017} and dice loss as adopted by our study}
\label{tab:8}       % Give a unique label
% For LaTeX tables use
\begin{tabular}{lll}
\hline\noalign{\smallskip}
loss & IoU & Macro F1-score \\
\noalign{\smallskip}\hline\noalign{\smallskip}
balanced cross-entropy & 0.8906 & 98.34\% \\
dice & 0.9144 & 99.21\% \\

\noalign{\smallskip}\hline
\end{tabular}
\end{table}
\section{Conclusion}
\label{conclusion}
In this work, we presented MRZNet, a framework specifically designed for localizing and recognizing the MRZ text in passport and visa images. A novel two stage model process is adopted so that MRZNet can handle passport/visa images of varies sizes from high resolution images.  We proposed MRZSpotter, an end-to-end network for detecting and recognizing MRZ text. By stacking multiple layers of ASPP, we increased the receptive field of the model and improved the MRZ text recognition accuracy. Experiment evaluation demonstrated the effectiveness of our approach compared with existing state-of-the-art models. Possible future research directions could include: 1) adding a dewarp component to the framework to make the pipeline robust to passport images that are warped with curved text lines; 2) modifying the architecture for single character level bounding box detection and recognition in order to further improve the overall robustness of the pipeline; 3) evaluating the performance of our models on passport/visa having MRZ region soiled by smoke, water/mud, ink or other artifacts.

\section{Declaration}
\subsection{Funding}
This research was supported by Lendbuzz.
\subsection{Conflict of Interest}
The authors declare that they have no conflict of interest.
\subsection{Availability of data and material}
Not available.
\subsection{Code availability}
Not available.
%\begin{acknowledgements}
%If you'd like to thank anyone, place your comments here
%and remove the percent signs.
%\end{acknowledgements}

% Authors must disclose all relationships or interests that 
% could have direct or potential influence or impart bias on 
% the work: 
%
% \section*{Conflict of interest}
%
% The authors declare that they have no conflict of interest.

% BibTeX users please use one of
% \bibliographystyle{spbasic}      % basic style, author-year citations
\bibliographystyle{spmpsci}      % mathematics and physical sciences
\bibliography{mrz_bibtex.bib}   % name your BibTeX data base

\section*{Author Biographies}
\setlength\intextsep{10pt}
  \vspace{-6pt}
\begin{wrapfigure}{r}{0.17\textwidth}
  \vspace{-15pt}
  \begin{center}
    \includegraphics[width=0.17\textwidth]{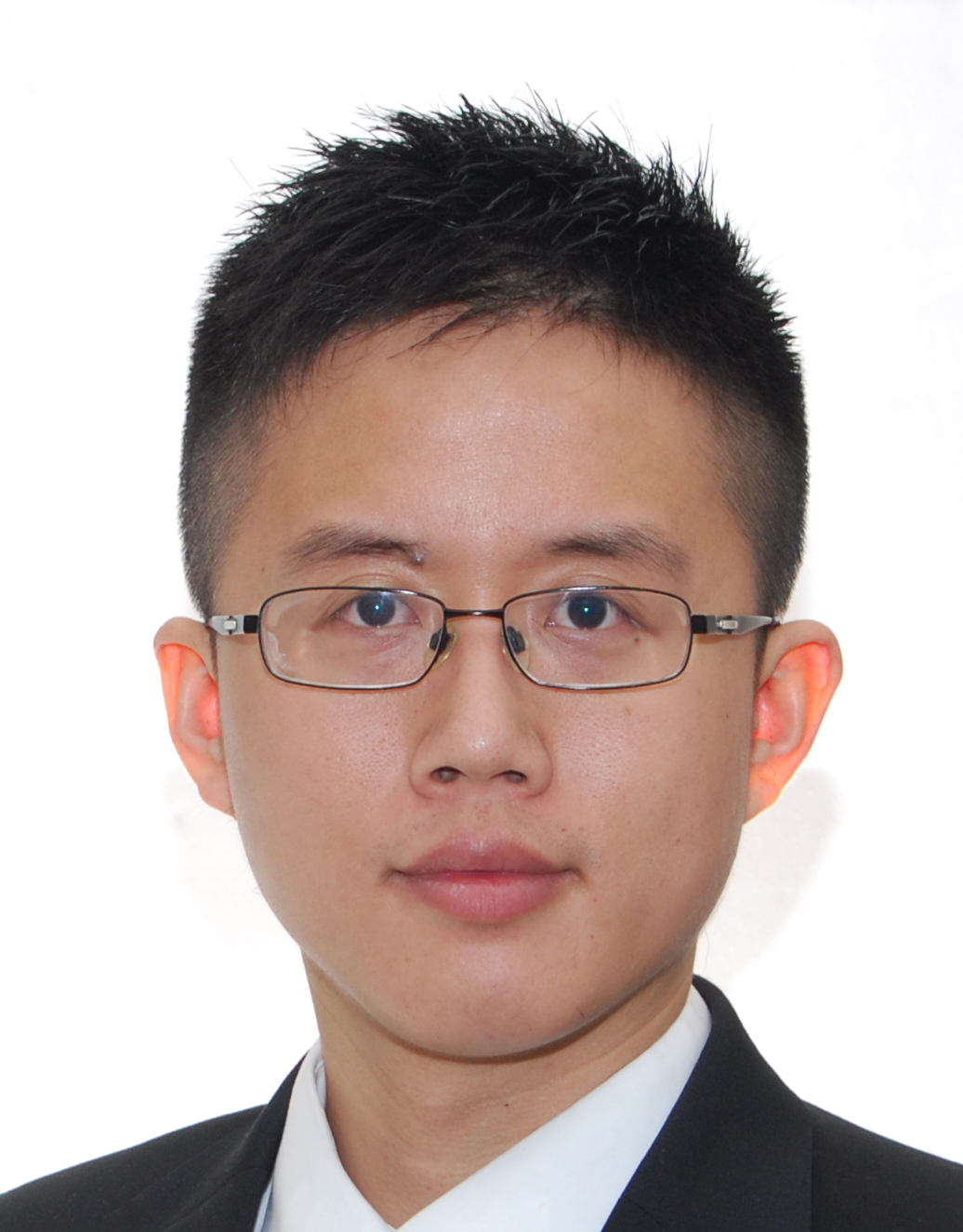}
  \end{center}
  \vspace{-10pt}
\end{wrapfigure}

\textbf{Yichuan} received Ph.D. degree in Biomedical Engineering from Drexel University in 2018. He is currently working as a machine learning engineer at Lendbuzz in Boston, Massachusetts, where his research includes document information extraction and document quality estimation and correction. Before joining Lendbuzz, he worked at Harvard Medical School where he developed deep learning algorithms to classify and segment medical images.\\

  \vspace{5pt}
\setlength\intextsep{10pt}
  \vspace{-6pt}
\begin{wrapfigure}{r}{0.17\textwidth}
  \vspace{-15pt}
  \begin{center}
    \includegraphics[width=0.17\textwidth]{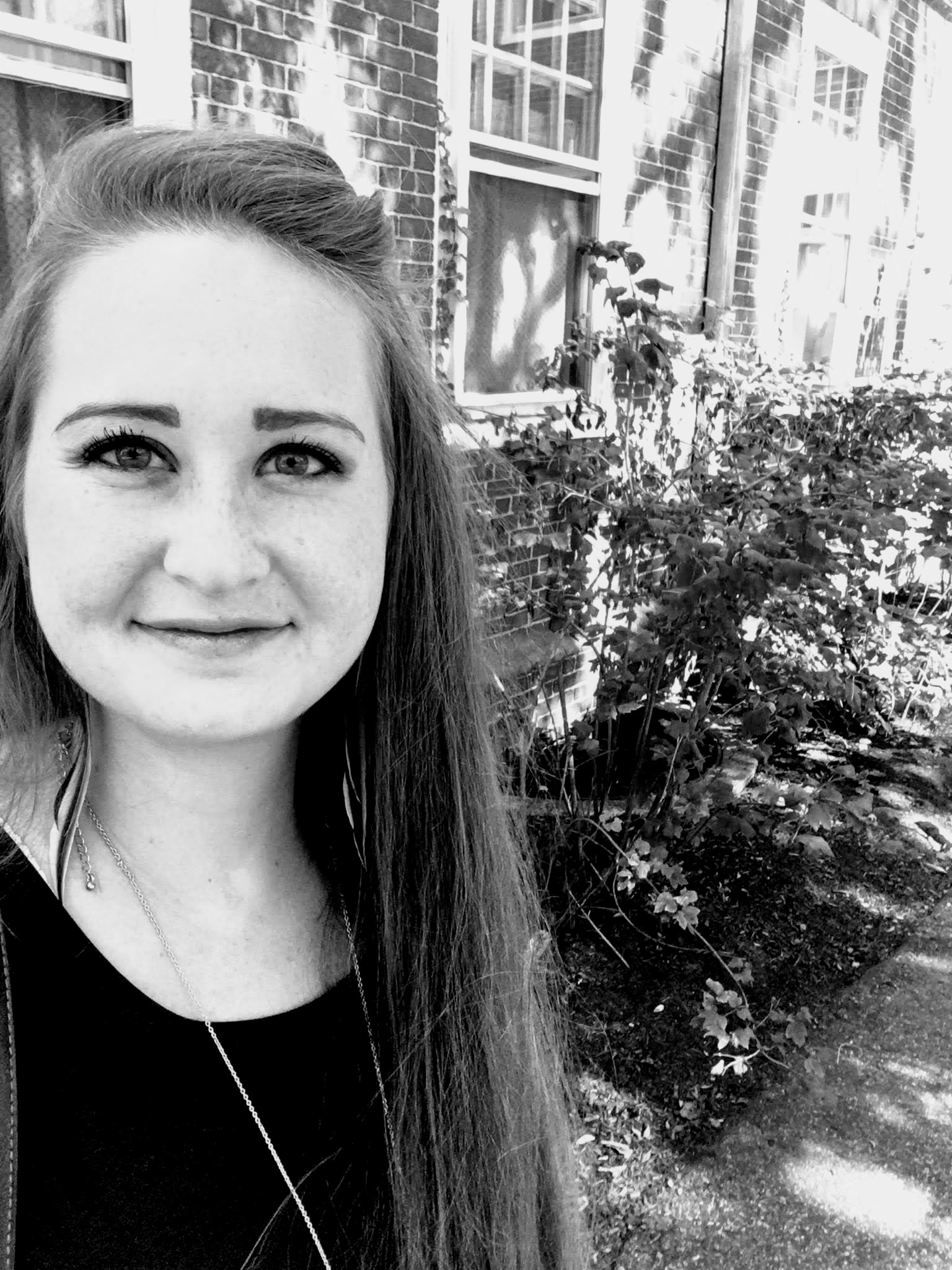}
  \end{center}
  \vspace{-10pt}
\end{wrapfigure}

\textbf{Hailey James} received Bachelor’s degree in Computer Science from Harvard College in Cambridge, Massachusetts in 2018. She is currently working as a machine learning engineer at Lendbuzz in Boston, Massachusetts, where her research includes image and document forensics. Her interests also include work in fairness, human-computer cooperation, and explainability in artificial intelligence.\\

  \vspace{5pt}
\setlength\intextsep{10pt}
\begin{wrapfigure}{r}{0.17\textwidth}
  \vspace{-15pt}
  \begin{center}
    \includegraphics[width=0.17\textwidth]{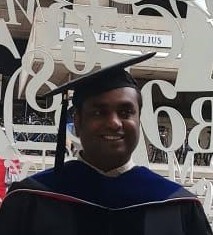}
  \end{center}
  \vspace{-10pt}
\end{wrapfigure}

\textbf{Otkrist Gupta} is currently Vice President of Data Science at Lendbuzz, focussing on deep learning with applications in finance and computer vision. He completed his Ph.D. at MIT Media Lab from camera culture group. His research is focused on inventing new algorithms for deep learning for health screening and diagnosis, hidden geometry detection, exploiting techniques from optimization, linear algebra and compressive sensing. He also works on designing algorithms for futuristic 3D projective displays.
Before joining MIT Media Lab Otkrist worked in Google Now team where he built voice actions such as take a picture and what’s on my Chromecast and worked on voice response quality from Google Now. He also worked at LinkedIn where he developed services such as Smart ToDo, Ultra fast auto-complete, Notifications and CheckIn platform. He completed his bachelors from Indian Institute of Technology Delhi (IITD) in Computer Science with emphasis on algorithms and linear algebra. After graduating from IITD, he worked for one year in field of High Frequency Trading at Tower Research Capital.\\

  \vspace{5pt}
\setlength\intextsep{10pt}
\begin{wrapfigure}{r}{0.17\textwidth}
  \vspace{-15pt}
  \begin{center}
    \includegraphics[width=0.17\textwidth]{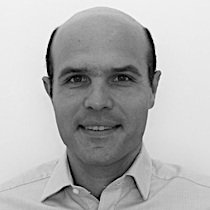}
  \end{center}
  \vspace{-10pt}
\end{wrapfigure}

\textbf{Dan Raviv} is the Chief Technology Officer of Lenbuzz Inc, an AI-based fintech underwriting company, and is faculty in the department of Engineering, Tel Aviv University, Israel. Dan did his post-doc at MIT in the Media Lab, working on various problems in the intersection of Geometry, Computer vision, and Machine learning, and is one of the leading researchers in Geometric Deep Learning. Dan was awarded the 2016 biennial award for Imaging Sciences granted by SIAM and published dozens of academic papers in his research field. Dan earned his Ph.D. and M.Sc. in computer science from the Technion – Israel Institute of Technology, Israel, and holds a bachelor’s degree in Mathematics and Computer Science, Summa Cum Lauda, from the Technion as well. Dan was a member of the Technion’s Excellence program, which handpicks the best and the brightest.\\

\end{document}